\documentclass{article}
\usepackage{spconf,amsmath,graphicx}
\usepackage{booktabs} 
\usepackage{graphicx} 

\usepackage{enumitem}

\title{Dual-Process Scaffold Reasoning for Enhancing LLM Code Debugging}

\name{Po-Chung Hsieh$^{1}$ \qquad Chin-Po Chen$^{2}$ \qquad Jeng-Lin Li$^{2}$ \qquad Ming-Ching Chang$^{3}$}
  
\address{$^{1}$National Taiwan University, Taipei, Taiwan \\
$^{2}$AI Research Center, Inventec Corporation \\
$^{3}$University at Albany, SUNY, Albany, NY, USA }

\begin{document}
\ninept
\maketitle

\begin{abstract}
Recent LLMs have demonstrated sophisticated problem-solving capabilities on various benchmarks through advanced reasoning algorithms. However, the key research question of identifying reasoning steps that balance complexity and computational efficiency remains unsolved. Recent research has increasingly drawn upon psychological theories to explore strategies for optimizing cognitive pathways. The LLM's final outputs and intermediate steps are regarded as System 1 and System 2, respectively. However, an in-depth exploration of the System 2 reasoning is still lacking. Therefore, we propose a novel psychologically backed Scaffold Reasoning framework for code debugging, which encompasses the Scaffold Stream, Analytic Stream, and Integration Stream. The construction of reference code within the Scaffold Stream is integrated with the buggy code analysis results produced by the Analytic Stream through the Integration Stream. Our framework achieves an 88.91\% pass rate and an average inference time of 5.36 seconds per-problem on DebugBench, outperforming other reasoning approaches across various LLMs in both reasoning accuracy and efficiency. Further analyses elucidate the advantages and limitations of various cognitive pathways across varying problem difficulties and bug types. Our findings also corroborate the alignment of the proposed Scaffold Reasoning framework with human cognitive processes.

\end{abstract}

\begin{keywords}
Large Language Models, LLM reasoning, code debugging, cognitive process.
\end{keywords}

\begin{figure*}[t]
\centerline{
  \includegraphics[width=\textwidth]{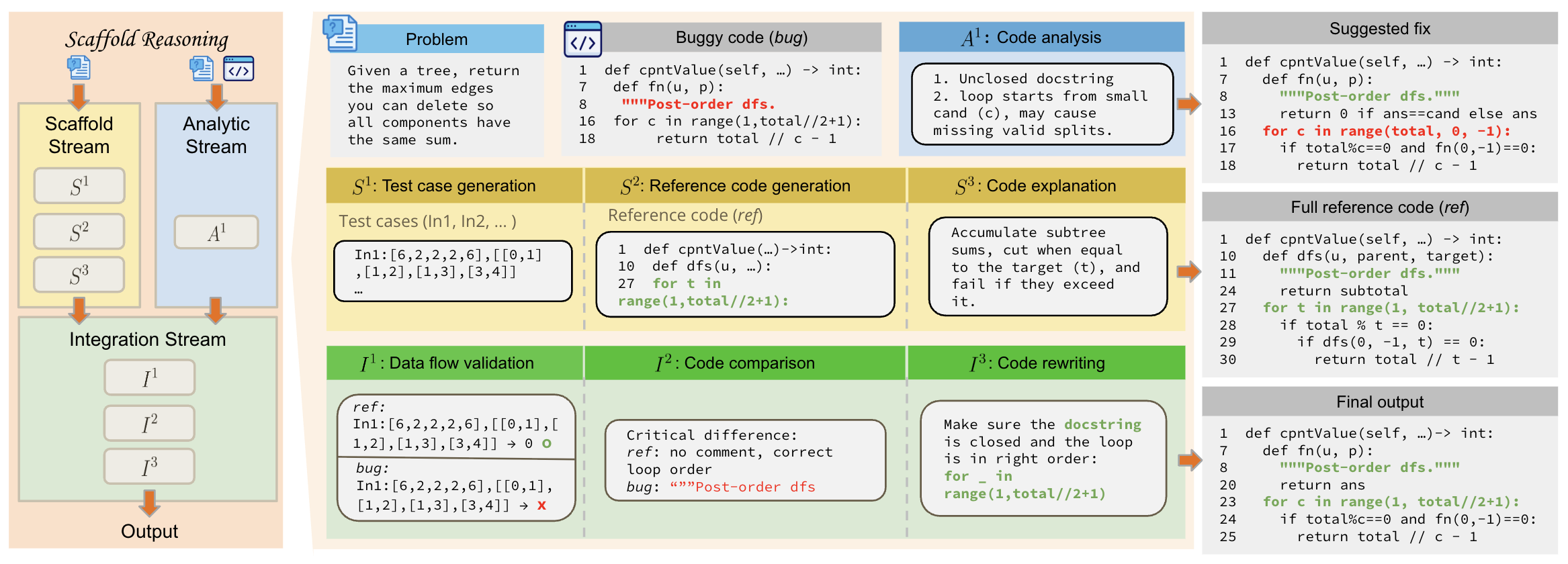} 
  \vspace{-3mm}
}  
\caption{\textbf{Scaffold Reasoning (SR) Framework} applied to the problem of {\em create-components-with-same-value}: (left) SR reasoning steps; (middle) a simplified code reasoning process illustrating each function of the framework. The Scaffold Stream produces reference code that is distinct yet logically aligned, supporting the Integration Stream when compared with the buggy code.
}
\label{fig:scaffold_reasoning}
\vspace{-3mm}
\end{figure*}

\section{Introduction}
\label{sec:intro}



Recent LLMs have shown improved reasoning across benchmarks, with methods such as multi-agent discussion, contextual guidance, and demonstration selection~\cite{zhang2025enhancing} boosting complex problem-solving. The common first step is planning, which restructures a problem into smaller subproblems, hypotheses, or statements. This is followed by iteratively validating the outcomes for further refinement~\cite{madaan2023selfrefine} and reflection~\cite{lee2025evolving}. This reasoning scheme has recently been extended to the multi-agent collaboration paradigm. Several prominent examples, including ReAct~\cite{yao2023reactsynergizingreasoningacting}, Chain-of-Agents~\cite{zhang2024chain}, and Chain-of-Unconscious-Thought~\cite {gong2025efficientreasoningchainunconscious}, increasingly mirror human reasoning, but they incur higher inference and communication costs, especially when generating and conveying interpretable reasoning steps. The challenge of designing reasoning pathways that balance reasoning complexity with computational efficiency remains unresolved.



Few studies have investigated the optimization of reasoning pathways, which is ideally inspired by psychological cognitive theories, such as mega-cognition and dual-process cognition theory. Dual-process theory differentiates between the rapid, intuitive operations of System 1 and the slower, deliberative reasoning of System 2. Current studies often oversimplify this by labeling direct LLM outputs as System 1 and any intermediate steps as System 2~\cite{yu2distilling}, without examining the nuanced reasoning within System 2. An open research question is which cognitive components and how their organization drive the most effective LLM reasoning. Deeper analysis is also needed to understand whether humans and LLMs share the same thinking pathways. 
Nevertheless, the core structure of reasoning pathways remains contested in various reasoning contexts. Some studies suggest increasing reasoning steps and cross-agent discussion can enhance response quality, even if the intermediate steps are not fully correct~\cite{jin2024impact}. However, others advocate limiting response length~\cite{li2025thinkless} and internalizing intermediate steps to attain an improved problem-solving success rate~\cite{gong2025efficientreasoningchainunconscious}. 
We aim to investigate the reasoning problem to reveal how the thinking pathways of System 2 more effectively address the code debugging problem.



Code debugging involves an act of reasoned communication where a developer interprets signals from the code and reconstructs the correct intended message. Recent attempts to simulate the human debugging process by applying the cognitive processing pipeline have improved fault localization and code fixing outcomes~\cite{zhong2024debuglikehumanlarge}. Strategically uncovering the underlying logic is crucial for sound reasoning, as demonstrated by the gains from data flow extraction in coding problems~\cite{wang-etal-2024-sanitizing}.
Although abstract-level concepts determine the reasoning direction, the complete answer still depends heavily on context~\cite{hua2025disentangling}. To the best of our knowledge, there has been limited exploration of structured thinking streams for uncovering the core steps of problem solving. Inspired by {\em mental scaffolding theory}~\cite{vygotsky1978mind}, which originally describes the temporary support provided to learners to help them achieve tasks they cannot complete independently, we construct a scaffolding thinking stream to generate pseudo-code and combine the analyzed bug localization thoughts to enhance the final fixing outcomes.


In this work, we propose a model-agnostic Scaffold Reasoning framework (Figure~\ref{fig:scaffold_reasoning}) for code debugging that consolidates two thinking pathways in a joint inference-time reasoning enhancement strategy. Our Scaffold Reasoning framework integrates an Analytic Stream alongside the Scaffold Stream to form a joint reasoning stream that consolidates both direct analysis and abstract guidance. This multi-stream architecture reflects a decomposed view of System 2 cognitive processes, illustrating how latent structures of thought are organized and operationalized for code debugging. Our experiments on DebugBench demonstrate a state-of-the-art reasoning pass rate of 88.91\%. We further perform an ablation study on the reasoning components to highlight the contribution of structured cognitive processes in addressing tasks with high cognitive complexity. 

Our contribution is summarized as follows:
\begin{itemize}[leftmargin=10pt] \itemsep -.1em

\item We introduce the \textbf{Scaffold Reasoning (SR) Framework}, which adapts cognitive scaffold theory to software engineering to enhance LLM-based code debugging. The framework structures reasoning into complementary streams, guiding models through both abstract and analytic problem-solving. Code will be made publicly available upon paper acceptance~\footnote{\scriptsize{Repo: https://github.com/scaffold-reasoning}.}.

\item Our SR achieves state-of-the-art pass rate of 88.91\% on DebugBench, demonstrating superior reasoning accuracy and efficiency compared to existing LLM reasoning methods.

\item We analyze the key cognitive steps that most influence reasoning performance, revealing how structured thinking pathways guide LLMs through complex problem-solving and substantially enhance debugging accuracy and efficiency.
    
\end{itemize}


\section{METHOD}

\subsection{Scaffold Reasoning Framework}

Our Scaffold Reasoning (SR) framework introduces a structured approach to LLM-based code debugging by organizing reasoning into three complementary streams: \textit{Scaffold Stream}, \textit{Analytic Stream}, and \textit{Integration Stream}. As shown in Figure~\ref{fig:scaffold_reasoning}, the Scaffold Stream builds a high-level understanding from the problem description, while the Analytic Stream inspects both the problem and buggy code to locate errors and propose fixes. Their outputs are then combined in the Integration Stream, which validates, compares, and synthesizes results into the final corrected code. This design reflects a decomposed view of System 2 cognition, separating abstract guidance from direct analysis before reconciling them for reliable debugging. We detail each stream below.

The \textbf{Scaffold Stream} consolidates an intuitive solution to the input problem, preserving the LLM’s natural problem-solving logic while promoting structured abstraction. To ensure high-quality code generation, it follows three key steps:
\begin{enumerate}[leftmargin=10pt,noitemsep,topsep=0pt]
\item Test case generation ($S^1$) automatically creates test cases for the given problem, serving as a validation suite for both the Analytic and Scaffold Streams and enhancing the reliability of subsequent debugging~\cite{lin-etal-2025-reasoning}.

\item Reference code generation ($S^2$) creates a reference implementation from the problem description, deliberately ignoring the buggy code and providing a clean conceptual baseline for later comparison.

\item Code explanation ($S^3$) generates a natural-language explanation of the reference code’s logic, encouraging  reflective reasoning and skin to self-reflection techniques shown to improve LLM performance~\cite{becker2024cyclesthoughtmeasuringllm}.

\end{enumerate}
Together, these steps scaffold an abstract reasoning pathway that anchors subsequent debugging to problem-level understanding rather than solely code-level heuristics.

\begin{figure}[t] 
\centerline{
  \includegraphics[width=\columnwidth]{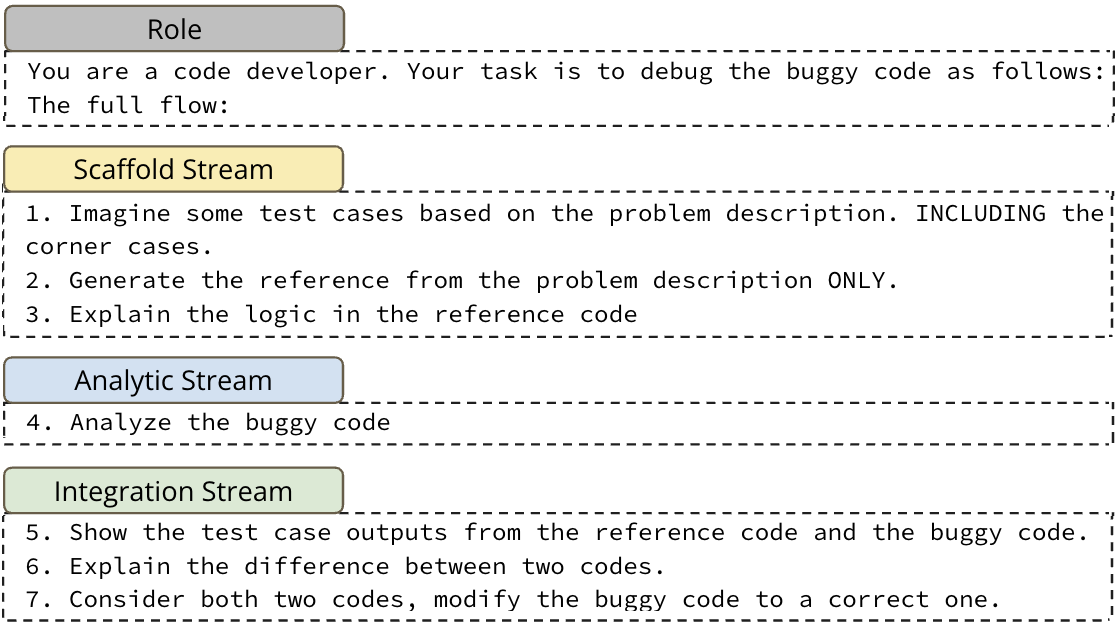}
  \vspace{-3mm}
}  
\caption{Prompts used in the Scaffold Reasoning Framework.}
\label{fig:prompt}
\vspace{-3mm}
\end{figure}

\begin{table*}[t]
\caption{Pass Rates of reasoning methods on DebugBench (\%). Bold indicates the best result. AvgPTime: Average Time Per Problem.}
\label{tab:main_result}
\centerline{
\footnotesize
\begin{tabular}{l|cccc|cc}
\toprule 
\textbf{Method} & \textbf{GPT-4o} & \textbf{GPT-4.1-mini} & \textbf{Devstral-Small-1.1} & \textbf{CodeQwen2.5-32B} & \textbf{Avg. Pass Rate (\%)↑} & \textbf{AvgPTime (s)↓} \\ 
\midrule 
\textbf{Base} & 84.33 & 85.56 & 67.80 & 84.04 & 80.43 & 5.62 \\ 
\textbf{CoT~\cite{wei2022chain}} & 84.22 & 85.21 & 67.61 & 83.80 & 80.21 & 6.93 \\ 
\textbf{Pearl~\cite{sun2024pearl}} & 85.25 & 84.59 & 66.61 & 82.16 & 79.65 & 7.17 \\ 
\textbf{ReAct~\cite{yao2023reactsynergizingreasoningacting}} & 85.65 & 86.40 & 67.90 & 83.69 & 80.91 & 5.49 \\ 
\textbf{LDB~\cite{zhong2024debuglikehumanlarge}} & 86.97 & 86.43 & 67.81 & 84.79 & 81.50 & 5.47 \\ 
\textbf{CoA~\cite{zhang2024chain}} & 84.97 & 83.50 & 67.27 & 77.91 & 78.41 & 6.36 \\ 
\midrule 
\textbf{SR} & \textbf{87.23} & \textbf{88.91} & \textbf{69.52} & \textbf{85.69} & \textbf{82.84} & \textbf{5.36} \\ 
\bottomrule
\end{tabular}
}
\vspace{-3mm}
\end{table*}

The \textbf{Analytic Stream} inspects the user’s buggy code to localize faults, analyze their root causes, and propose candidate fixes. We integrate bug localization techniques~\cite{saha2013improving} into this stream via Code analysis ($A^1$), which identifies faulty code regions and provides analytic descriptions of the issues while preserving logical coherence after modification. This stream offers a bottom-up reasoning pathway grounded in the code itself, complementing the abstract guidance of the Scaffold Stream.




The \textbf{Integration Stream} reconciles the intermediate outputs of the Scaffold and Analytic Streams, ensuring correctness, alignment, and synthesis into a reliable final solution. It operates in three steps:
\begin{enumerate}[leftmargin=10pt] \itemsep -.1em

\item Data flow validation ($I^1$) executes both candidate codes on the test cases generated by $S^1$ to verify correctness and detect inconsistencies.

\item Code comparison ($I^2$) examines logical alignment and line-by-line differences between the two streams’ outputs.

\item Code rewriting ($I^3$) generates the final corrected solution by integrating validated insights, balancing analytic bug fixes with abstract structural guidance.

\end{enumerate}
This structured integration enforces consistency between abstraction and analysis, reducing the risk of overfitting to either incomplete high-level reasoning or narrow bug-fix heuristics.

\subsection{Prompt Implementation}

To illustrate the SR framework, we provide a worked example using the problem {\em create-components-with-same-value}. The framework processes the problem and buggy code as follows:
\begin{enumerate}[leftmargin=10pt] \itemsep -.1em

\item Analytic Stream ($A^1$) localizes two issues: (i) an unclosed docstring in \texttt{fn}, and (ii) an incorrect loop iterating from $\texttt{1}$ to $\texttt{total//2+1}$, which risks missing the largest valid splits. It suggests closing the docstring and adjusting the loop to iterate from $\texttt{total}$ down to $\texttt{1}$.

\item Scaffold Stream $S^1$ generates test cases covering diverse tree structures. $S^2$ produces reference code that iterates candidates from $\texttt{1}$ to $\texttt{total//2+1}$ and applies DFS to verify valid component splits. $S^3$ explains the reasoning by accumulating subtree sums, cutting edges when sums match the candidate, and propagating failure if sums exceed the candidate value.

\item Integration Stream $I^1$ validates both codes, showing that the buggy code produces no output due to the unclosed docstring, while the reference code is correct. $I^2$ compares the output, noting that relying solely on the Analytic Stream may introduce errors, such as incorrect loop order. $I^3$ synthesizes the final solution by closing the docstring and preserving the correct loop order, ensuring correctness and maximizing deletable edges.

\end{enumerate}
This example demonstrates how the SR framework decomposes reasoning into complementary streams and integrates them to produce robust debugging outcomes. Figure~\ref{fig:prompt} shows the prompt template implementing the SR framework.

\section{Experiments}


\noindent\textbf{Datasets:}
We evaluate our approach using DebugBench~\cite{tian2024debugbenchevaluatingdebuggingcapability}, a comprehensive benchmark that contains a large collection of buggy code examples to test the LLMs' debugging and code repair abilities. The correctness of repaired solutions can be assessed by submitting them to the LeetCode backend for validation. We focus on the Python subset of DebugBench, which contains 1,414 debugging questions. The longest buggy code spans 103 lines, while the average length is 22.98 lines with a standard deviation of 14.12. This subset includes diverse bug categories, including {\em syntax errors}, {\em logical errors}, {\em reference errors}, and {\em multiple errors} (731 cases). The coding problems are classified into three difficulty levels: {\em Easy}, {\em Medium}, and {\em Hard}.


\noindent\textbf{Implementation details:}
We applied different reasoning methods to the following models: \textbf{GPT-4.1-mini}, \textbf{GPT-4o}~\cite{openai2024gpt4technicalreport}, \textbf{Devstral-Small-1.1}~\footnote{\scriptsize{https://huggingface.co/mistralai/Devstral-Small-2507}}, and \textbf{CodeQwen2.5-32B}~\cite{hui2024qwen25codertechnicalreport}. The GPT-series models were queried using the Azure OpenAI API~\footnote{\scriptsize{https://portal.azure.com}}, while the open-source models were run on an Azure VM equipped with an NVIDIA H100 GPU. For each inference, we directly prompted the LLMs in a zero-shot manner.

\noindent\textbf{Baselines:}
We evaluate our approach against several established baselines, including \emph{Base}, \emph{PEARL}, \emph{ReAct}, \emph{LDB}, and \emph{CoA}, using the LeetCode API~\footnote{\scriptsize{https://leetcode.com}}. The evaluation metric, \textbf{Pass Rate}, is defined as the proportion of problems for which the code successfully passes all test cases divided by the total problem count. The details of each method are as follows:  
\begin{itemize}[leftmargin=10pt] \itemsep -.1em
    \item \textbf{Base}: directly prompts the LLM to correct buggy code without structured reasoning.  
    \item \textbf{CoT}~\cite{wei2022chain}: implements chain-of-thought by directly prompting the LLM to further reasoning on buggy code correction with ''thinking step-by-step''.
    \item \textbf{PEARL}~\cite{sun2024pearl}: decomposes debugging into multiple actions, such as summarize, locate error, suggest a fix, and so on. The LLM inference is performed in an action-by-action manner to derive the final solution.
    \item \textbf{ReAct}~\cite{yao2023reactsynergizingreasoningacting}: introduces a prompt-based paradigm that resembles human adaptive thinking by iteration over three thinking steps: \emph{thinking}, \emph{action}, and \emph{observation}. 
    \item \textbf{LDB}~\cite{zhong2024debuglikehumanlarge}: uses runtime execution traces and intermediate variables to resemble the debugging practices of human developers.  
    \item \textbf{CoA}~\cite{zhang2024chain}: proposes a multi-agent framework for code debugging, where each worker processes a portion of the program to extract key information, and their outputs are aggregated and organized by a manager agent to guide repair.
\end{itemize}

\begin{table*}[t]
\caption{Pass Rates and AvgPTime in the ablation study. Bold indicates the best performance in each column.}
\label{tab:ablation}
\centerline{
\footnotesize
\setlength{\tabcolsep}{1mm}
\begin{tabular}{l|cc|ccc|cccc}
\toprule
\textbf{} & \textbf{Pass Rate (\%)↑} & \textbf{Time (s) ↓} & \textbf{Easy (\%)↑} & \textbf{Medium (\%)↑} & \textbf{Hard (\%)↑} & \textbf{Syntax (\%)↑} & \textbf{Logic (\%)↑} & \textbf{Reference (\%)↑} & \textbf{Multiple (\%)↑} \\
\midrule
\textbf{SR}                             & \textbf{88.91} & 5.36 & 95.13 & 92.77 & \textbf{77.78} & \textbf{92.82} & \textbf{95.52} & 91.28 & 85.01 \\
\textbf{SR}-S                   & 86.70 & 6.20 & 92.64 & 92.59 & 75.14 & 90.39 & 87.90 & 89.50 & 84.19 \\
\textbf{SR}-A                   & 88.30 & 6.34 & \textbf{96.39} & \textbf{93.70} & 75.43 & 89.50 & 90.70 & \textbf{91.54} & \textbf{86.22} \\
\textbf{SR}-$I^2$+$I^{2*}$ & 87.07 & 5.77 & 94.00 & 90.16 & 77.50 & 92.23 & 88.96 & 90.06 & 83.80 \\
\textbf{SR}-$S^2$+$S^{2*}$            & 86.96 & \textbf{5.17} & 93.82 & 90.51 & 76.06 & 92.05 & 87.95 & 89.63 & 84.05 \\
\textbf{SR}-$S^1$-$S^3$              & 87.98 & 5.59 & 93.81 & 93.08 & 77.10 & 91.51 & 90.74 & 91.44 & 84.90 \\

\bottomrule
\end{tabular}
}
\vspace{-3mm}
\end{table*}




{\bf Experimental Results:}
Table~\ref{tab:main_result} shows the results of our Scaffold Reasoning (SR) framework compared to those of other recent studies. \textbf{SR} achieves 82.84\% Average Pass Rate, outperforming all baselines, such as \textbf{LDB} at 81.50\% and \textbf{ReAct} at 80.91\%. \textbf{Base} serves as a strong baseline with 80.43\% pass rate. The phenomenon that \textbf{CoT}, \textbf{PEARL}, and \textbf{CoA} fall below at 80.21\%, 79.65\%, and 78.41\% suggests the potential risk of overthinking.
In terms of Average Time Per Problem (AvgPTime), \textbf{SR} achieves the shortest 5.36 seconds by internalizing the thinking steps in a single inference template, compared to baselines ranging from 5.47 seconds (\textbf{LDB}) to 7.17 seconds (\textbf{PEARL}). Our proposed SR framework integrates dual thinking streams that consistently outperform all other methods, indicating the superiority of combining multi-directional reasoning over traditional linear thinking approaches.

{\bf Analysis of Results:}
Our Scaffold Reasoning framework meticulously orchestrates the logical flow and cognitive steps throughout the complete dual thinking stream. We decompose and reorganize the thinking steps to examine the ablation results. To further explore the differences between thinking streams, we present the pass rate of problems categorized by the problem difficulty (Easy, Medium, Hard) and bug types (Syntax, Logic, Reference, Multiple). These meta-categories are provided in the DebugBench dataset.




\subsection{Ablation Results of Reasoning Path and Reasoning Steps}

We conducted two ablation studies on the reasoning paths and  steps.
(1) The reasoning path ablation isolates a single stream of thought to assess its individual contribution. (2) The reasoning step ablation modifies or removes specific key steps to evaluate their impact. Scaffold Stream, Analytic Stream, and Integration Stream are denoted with $S$, $A$, and $I$, respectively. Each step numbers and its variance are denoted in the superscript, e.g., $S^{2*}$ denotes the modified $S$ in its second step. The following thinking streams are evaluated:
\begin{itemize}[leftmargin=10pt] \itemsep -.1em
    \item \textbf{SR}-S only preserves the Analytic Stream and then directly asks the LLM to fix the code.
    \item \textbf{SR}-A only preserves the Scaffold Stream and then directly asks the LLM to fix the code.
    \item \textbf{SR}-$I^2$+$I^{2*}$ constrains the code comparison step in the Integration Stream by considering common bug types, including logic, syntax, and condition ($I^{2*}$). 
    \item \textbf{SR}-$S^2$+$S^{2*}$ replaces the reference code generation step in the Scaffold Stream with pseudocode generation ($S^{2*}$).
    \item \textbf{SR}-$S^1$-$S^3$ removes both test case generation and code explanation steps in the Scaffold Stream.
\end{itemize}




The top three rows of Table~\ref{tab:ablation} show the reasoning path ablation results. Our \textbf{SR} achieves the highest pass rate (88.91\%) and the shortest processing time (5.36 seconds). 
Using only \textbf{SR}-S thinking stream, the pass rate drops to 86.70\% and solving time rises to 6.20 s, while \textbf{SR}-A yields a pass rate of 88.30\% with increased processing time of 6.34 s. These results indicate that relying solely on one stream (without a coordinated reasoning flow) reduces efficiency and effectiveness. The superior performance of the full \textbf{SR} framework confirms the value of the dual-stream design. Notably, the increased processing times for \textbf{SR}-S and \textbf{SR}-A suggest that structured thinking streams help LLMs reduce cognitive load by guiding them through the solution more efficiently.


The bottom three rows of Table~\ref{tab:ablation} show the reasoning step ablation results. Modifying the code comparison step in the Integration Stream (\textbf{SR}-$I^2$+$I^{2*}$) yields a pass rate of 87.07\% with an average solving time of 5.77 seconds. Compared to the automatic code comparison performed by LLMs, the human-designed scheme shows clear limitations.
Removing both test case generation and code explanation steps in the Scaffold Stream (\textbf{SR}-$S^1$-$S^3$) still maintains a relatively high pass rate of 87.97\% with 5.59 seconds per problem. In contrast, replacing reference code generation with pseudocode generation (\textbf{SR}-$S^2$+$S^{2*}$) results in a substantial decline in pass rate to 86.96\%. This comparison reveals that the reference code generation step $S^2$ plays the most critical role in the Scaffold Stream, while $S^1$ and $S^3$ provide auxiliary checking strategies for refinement. 
The generation of reference code parallels schema construction~\cite{mcvee2005schema}, initializing organization of concepts that support subsequent refinement and problem-solving. The schema functions as an internal representation within thinking steps of the mental scaffold, consistent with established psychological theories. Scaffolds generally serve as preliminary constructs built with past schema and are often less effective than the fully developed and refined solutions they precede.




\subsection{Reasoning Effects on Problem Difficulties and Bug Types}


We investigate the results of each problem subset categorized by problem difficulty (Easy, Medium, Hard) and buggy types (Syntax, Logic, Reference, Multiple). The middle columns of Table~\ref{tab:ablation} show these results. \textbf{SR}-A performs well on easy and medium problems with 96.39\% and 96.70\% pass rates, yet fails on hard ones, suggesting that LLMs can handle the easier problems without the Analytic Stream. When buggy code involves reference errors or multiple bug types, the inconsistency in logic makes it hard for the LLM to repair. Writing new code from scratch using the Scaffold Stream becomes a more effective strategy.
However, the code purely from Scaffold Stream still suffers from syntax and logic errors, performing 3.32\% and 4.82\% worse than \textbf{SR}. 
Finally, we observe that no single ablated thinking pathway closely matches \textbf{SR} in solving logic errors, due to the need for both structured reasoning and analytical verification to resolve flaws in underlying logic.

\section{Conclusion}

In this work, we leverage psychological theory to design the Scaffold Reasoning (SR) framework and its intermediate reasoning steps for code debugging. Our SR framework achieves a top pass rate of 88.91\% on Python coding tasks in DebugBench, outperforming other reasoning approaches across multiple LLMs. By examining diverse thinking pathways and identifying essential reasoning steps, we systematically validate optimal reasoning strategies for LLMs.
Our results confirm the effectiveness of the dual-stream design: the Scaffold Stream constructs reference code, the Analytic Stream performs in-depth analysis, and the Integration Stream refines the solution through multiple verification mechanisms. This architecture aligns closely with psychological models of human cognition. By internalizing the entire reasoning process within a single inference run, SR achieves both high accuracy and efficiency, demonstrating strong potential for practical applications.

{\bf Future work} includes integrating the dual-thinking-path design into model training and conducting human cognition studies to observe human developers’ debugging strategies. Given that LLMs can emulate human reasoning, these cognitively aligned designs may drive improvements across multiple domains, facilitating cross-disciplinary advancements.

\bibliographystyle{IEEEbib}
\bibliography{reasoning}

\end{document}